\documentclass[10pt,twocolumn,letterpaper]{article}

\usepackage[pagenumbers]{cvpr} %

\usepackage{algorithm}
\usepackage{algpseudocode}
\usepackage{booktabs}

\usepackage{multicol}
\usepackage[dvipsnames]{xcolor}

\usepackage{listings}
\usepackage{xcolor}
\usepackage{float}

\definecolor{codegreen}{rgb}{0,0.6,0}
\definecolor{codegray}{rgb}{0.5,0.5,0.5}
\definecolor{codepurple}{rgb}{0.58,0,0.82}
\definecolor{backcolour}{rgb}{0.95,0.95,0.92}

\lstdefinestyle{mystyle}{
    backgroundcolor=\color{backcolour},   
    commentstyle=\color{codegreen},
    keywordstyle=\color{magenta},
    numberstyle=\tiny\color{codegray},
    stringstyle=\color{codepurple},
    basicstyle=\ttfamily\footnotesize,
    breakatwhitespace=false,         
    breaklines=true,                 
    captionpos=b,                    
    keepspaces=true,                 
    numbers=left,                    
    numbersep=5pt,                  
    showspaces=false,                
    showstringspaces=false,
    showtabs=false,                  
    tabsize=2
}

\newfloat{lstfloat}{htbp}{lop}
\floatname{lstfloat}{Algorithm}

\lstdefinestyle{myverbatim}{
    basicstyle=\ttfamily\footnotesize,
    backgroundcolor=\color{white},
    breaklines=true,
    breakatwhitespace=true
}

\usepackage[para]{footmisc}
\usepackage{color}
\usepackage{bm}
\usepackage{placeins}
\usepackage{xcolor}
\usepackage{colortbl}
\usepackage{enumitem}
\usepackage{amsmath}
\usepackage{adjustbox}
\definecolor{cvprblue}{rgb}{0.21,0.49,0.74}
\definecolor{Green}{RGB}{0,155,85}
\definecolor{LightOrange}{rgb}{1,0.85,0.8}
\definecolor{LightPurple}{rgb}{1.0,0.80,0.95}
\definecolor{LightGreen}{rgb}{0.93,0.98,0.96}
\usepackage[pagebackref,breaklinks,colorlinks,citecolor=cvprblue]{hyperref}

\def\paperID{568}
\def\confName{CVPR}
\def\confYear{2024}

\title{Self-correcting LLM-controlled Diffusion Models}

\author{Tsung-Han Wu\footnotemark[1] \qquad Long Lian\footnotemark[1] \qquad Joseph E. Gonzalez \qquad Boyi Li\footnotemark[2] \qquad Trevor Darrell\footnotemark[2] \\
  UC Berkeley \\
}

\begin{document}

\twocolumn[{
\renewcommand\twocolumn[1][]{#1}
\maketitle
\begin{center}
    \centering
    \vspace{-2em}
    \includegraphics[width=\linewidth]{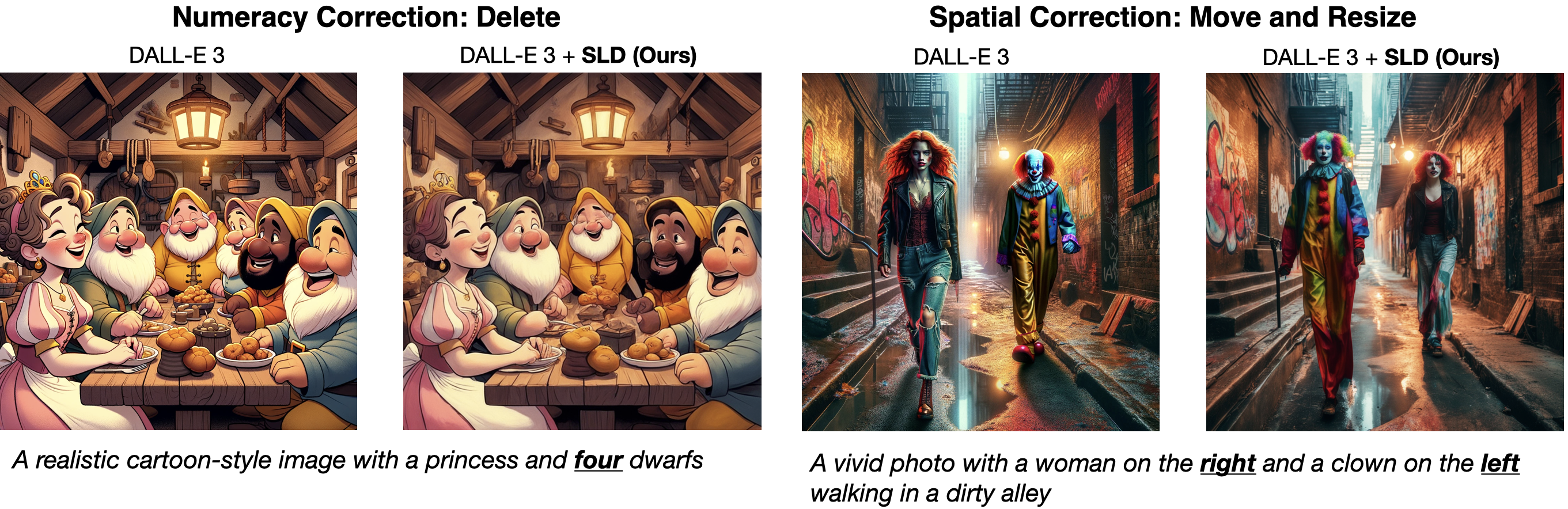}
    \vspace{-1em}
    \captionof{figure}{Existing diffusion-based text-to-image generators (e.g., DALL-E 3 \cite{dalle3}) generally struggle to precisely generate images that correctly align with complex input prompts, especially for the ones that require numeracy and spatial relationships. Our Self-correcting LLM-controlled Diffusion (SLD) framework empowers these diffusion models to automatically and iteratively rectify inaccuracies through applying a set of latent space operations (addition, deletion, repositioning, \etc), resulting in enhanced text-to-image alignment.}
    \label{fig:teaser}
\end{center}
}]

\def\figMain#1{
\begin{figure*}[#1]
\centering
\vspace{-8pt}
\includegraphics[width=0.9\linewidth]{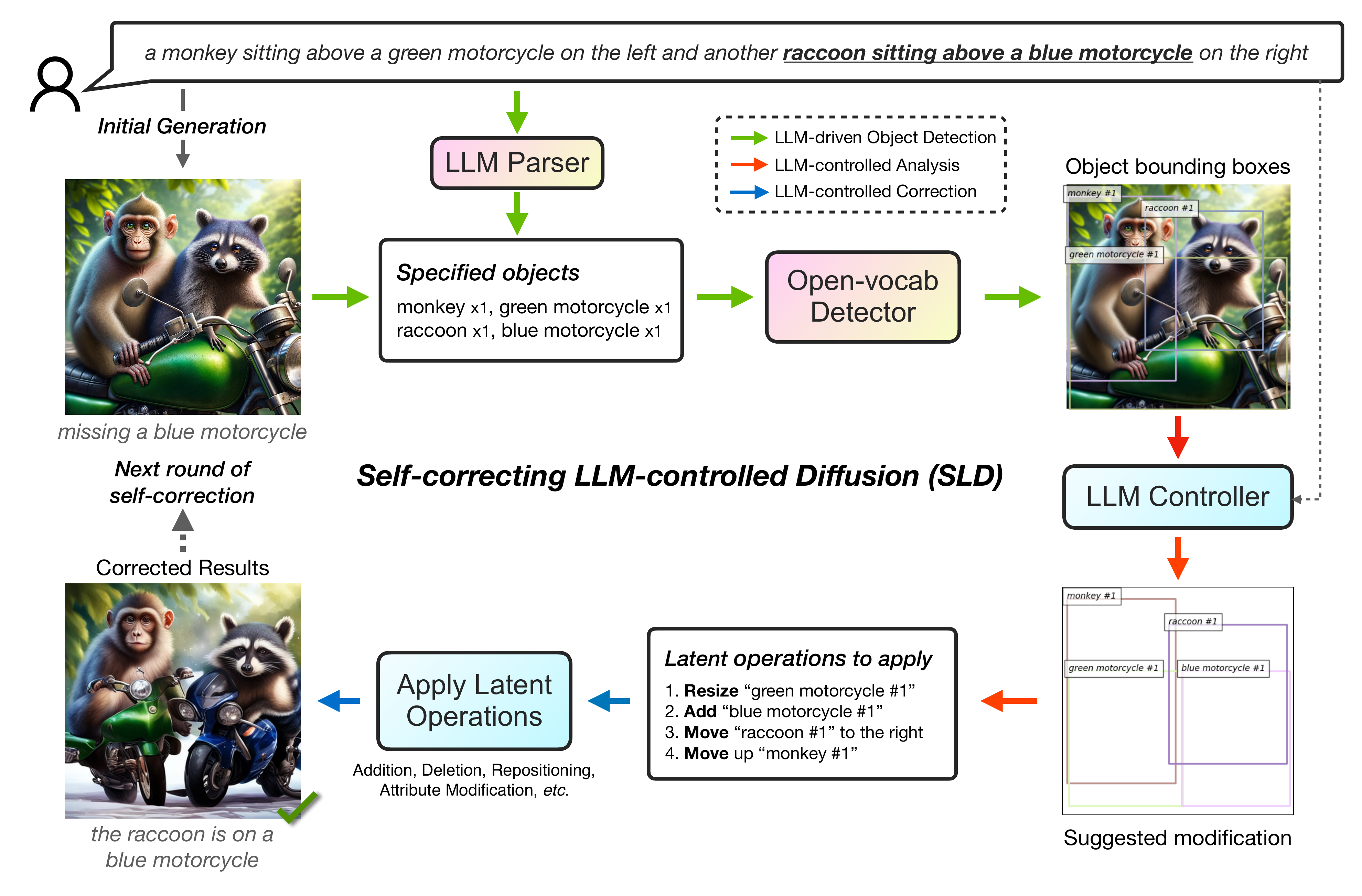}
\vspace{-12pt}
\caption{Our proposed Self-correcting LLM-controlled Diffusion (SLD) enhances text-to-image alignment through an iterative self-correction process. It begins with LLM-driven object detection (\cref{ssec:object_detection}), and subsequently performs LLM-controlled analysis and correction (\cref{ssec:llm_controlled_image_analysis}). The entire pipeline is outlined in \cref{alg:self-correct-image-generation}.}
\vspace{-6pt}
\label{fig:main}
\end{figure*}
}

\def\figPrompt#1{
\begin{figure*}[#1]
\centering
\includegraphics[width=1.0\linewidth]{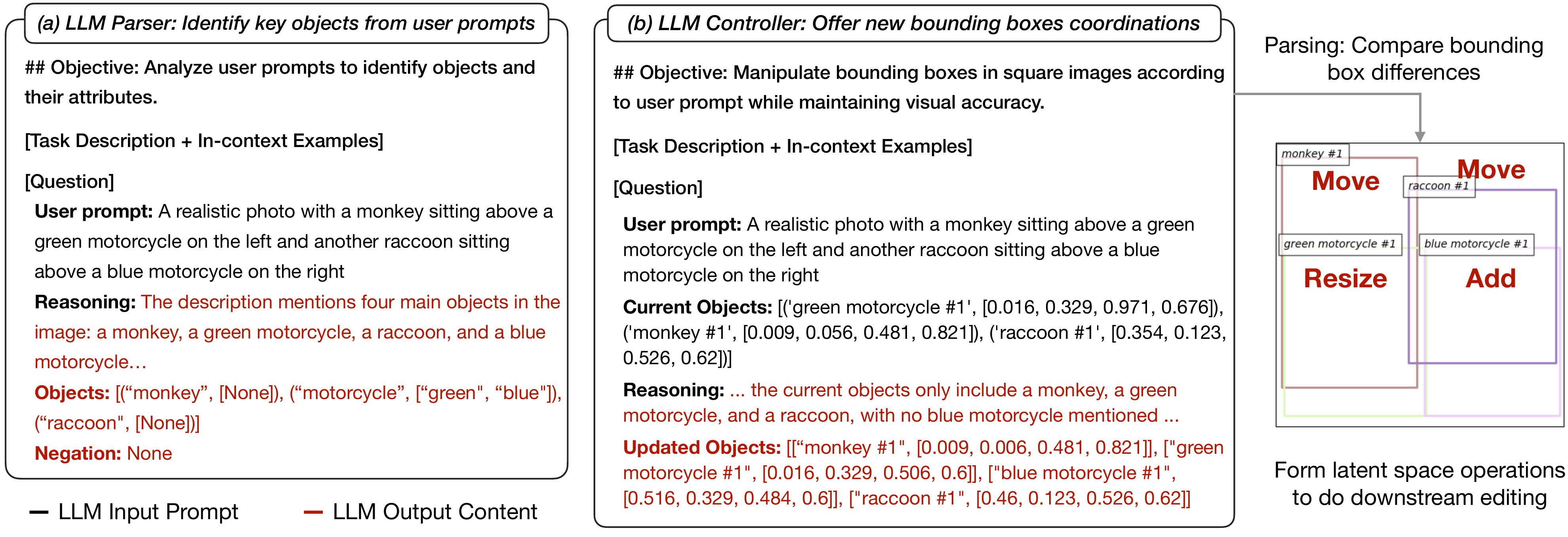}
\vspace{-16pt}
\caption{Our self-correction pipeline is driven by two distinct LLMs: \textbf{(a)} The LLM parser analyzes user prompts $P$ to extract a list of key object information $S$, which is then passed to the open-vocabulary detector. \textbf{(b)} The LLM controller, taking both the user prompt $P$ and currently detected bounding boxes $B_{curr}$ as input, outputs suggested new bounding boxes $B_{next}$. These are subsequently transformed into a set of latent space operations $Ops$ for image manipulation.}
\vspace{-6pt}
\label{fig:prompt}
\end{figure*}
}

\def\figLatent#1{
\begin{figure}[#1]
\centering
\includegraphics[width=1.0\linewidth]{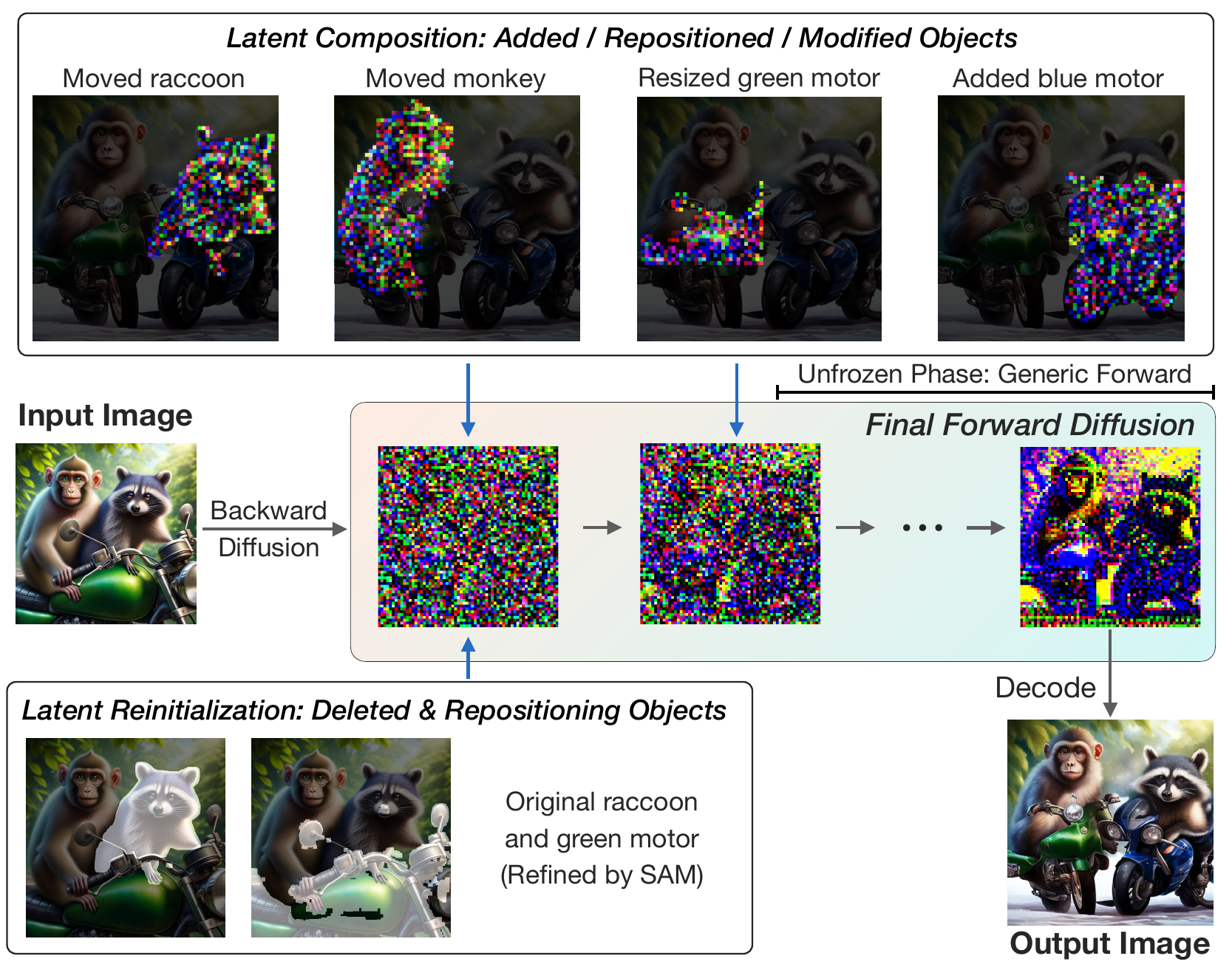}
\vspace{-16pt}
\caption{Our latent operations can be summarized into two key concepts: (1) latent in removed regions are re-initialized to Gaussian noise, and latent of newly added or modified objects are composited onto the canvas. (2) Latent composition is confined to the initial steps, followed by ``unfrozen" steps for a standard forward diffusion process, enhancing visual quality and avoiding artificial copy-and-paste effects.}
\vspace{-6pt}
\label{fig:latent}
\end{figure}
}

\def\figVis#1{
\begin{figure*}[#1]
\centering
\includegraphics[width=1.0\linewidth]{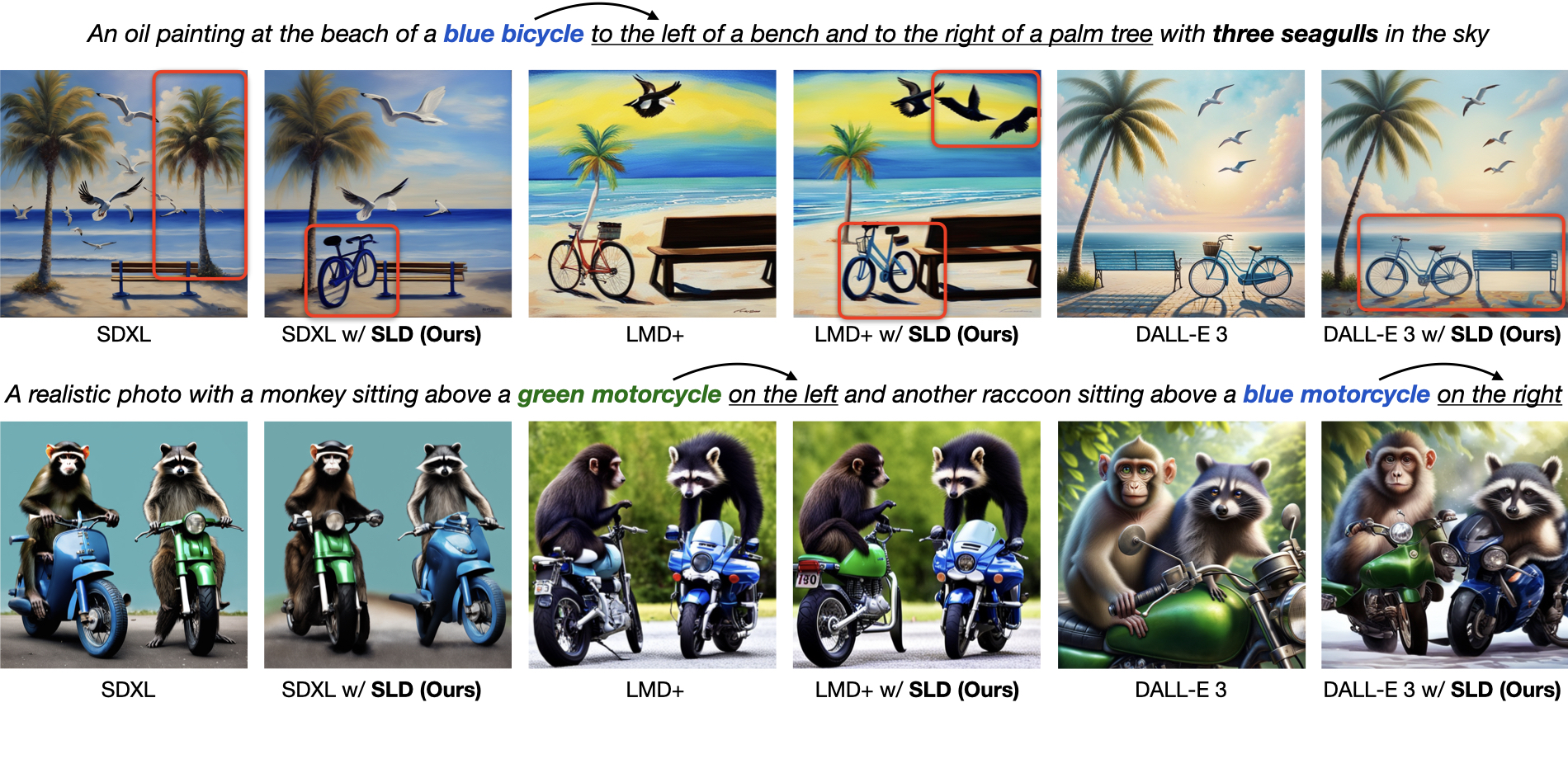}
\vspace{-32pt}
\caption{SLD enhances text-to-image alignment across diverse diffusion-based generative models such as SDXL, LMD+, and DALL-E 3. Notably, as highlighted by the red boxes in the first row, SLD precisely positions a blue bicycle in relation to a bench and a palm tree, while maintaining the accurate count of palm trees and seagulls. The second row further demonstrates SLD's robustness in complex, cluttered scenes, effectively managing object collision through our training-free latent operations.}
\label{fig:vis}
\end{figure*}
}

\def\figEdit#1{
\begin{figure*}[#1]
\centering
\includegraphics[width=1.0\linewidth]{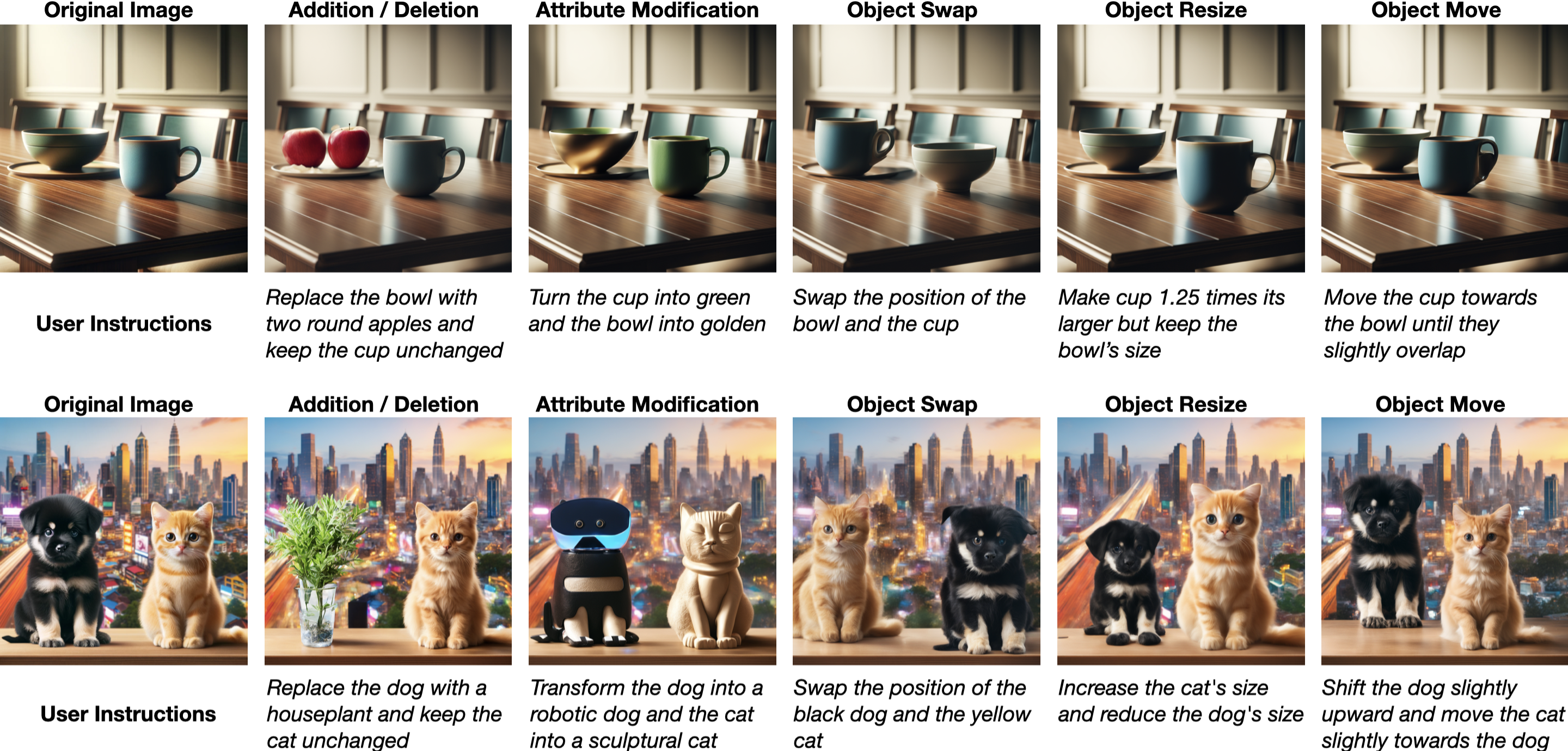}
\vspace{-8pt}
\caption{SLD can handle a diverse array of image editing tasks guided by natural, human-like instructions. Its capabilities span from adjusting object counts to altering object attributes, positions, and sizes.}
\vspace{-6pt}
\label{fig:edit}
\end{figure*}
}

\def\Algo#1{
\begin{algorithm}[#1]
\caption{Self-correction for Image Generation.}
\label{alg:self-correct-image-generation}
\begin{algorithmic}[1]
\Require User prompt $P$, Initial generated image $I$, Maximum number of self-correction round $K$.

\For{$k \leftarrow 1$ to $K$}

\State $S \leftarrow \texttt{LLM-Parser}(P)$
\State $B_{curr} \leftarrow \texttt{Detector}(S)$
\State $B_{next} \leftarrow \texttt{LLM-Analysis}(P, B_{curr})$
\State $Ops \leftarrow \texttt{Diff}(B_{curr}, B_{next})$
\If{$Ops \neq \emptyset$} 
(i.e., $B_{next} \neq B_{curr}$)
\State $I = \texttt{Correction}(Ops, B_{next}, B_{curr})$
\Else
\State \textbf{break}
\EndIf{}
\EndFor{}
\Ensure Image $I$.
\end{algorithmic}
\end{algorithm}
}

\def\figEditCompare#1{
\begin{figure}[#1]
\centering
\vspace{-8pt}
\includegraphics[width=0.9\linewidth]{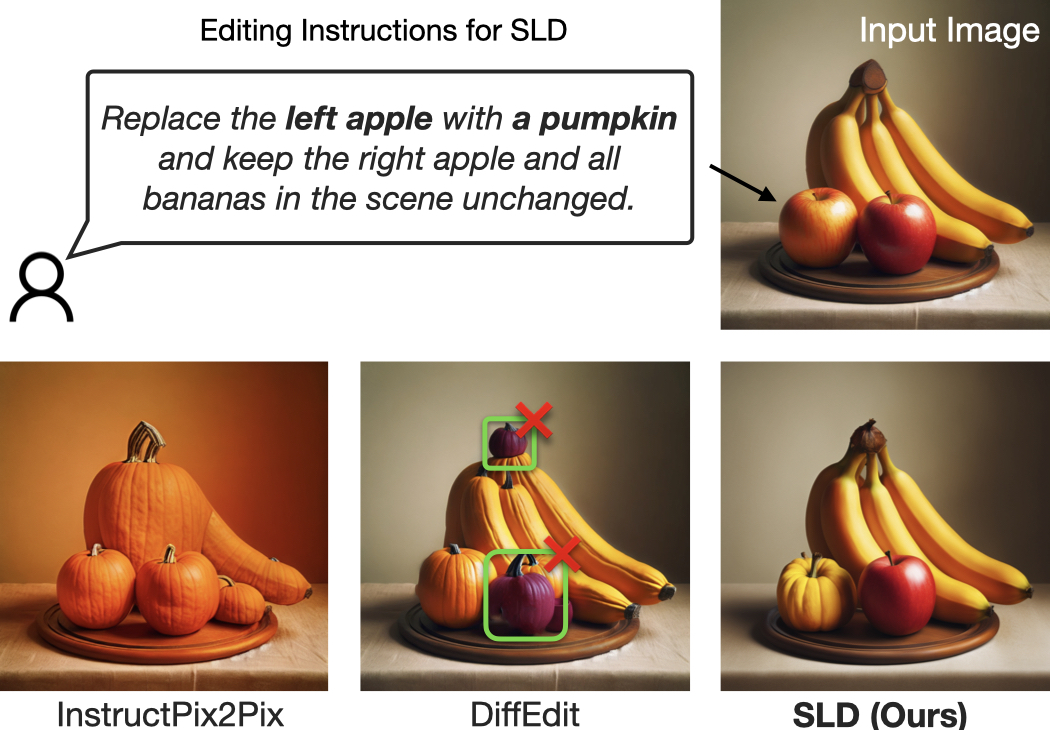}
\vspace{-4pt}
\caption{When instructed to perform object-level image editing. InstructPix2Pix \cite{brooks2023instructpix2pix} completely fails to accomplish the task, and DiffEdit \cite{couairon2023diffedit} falls short, as highlighted in the green box of the image. Conversely, our method demonstrates a significantly better performance in executing these object-level edits.}
\vspace{-6pt}
\label{fig:edit_comparison}
\end{figure}
}

\def\figFailure#1{
\begin{figure}[#1]
\centering
\includegraphics[width=1.0\linewidth]{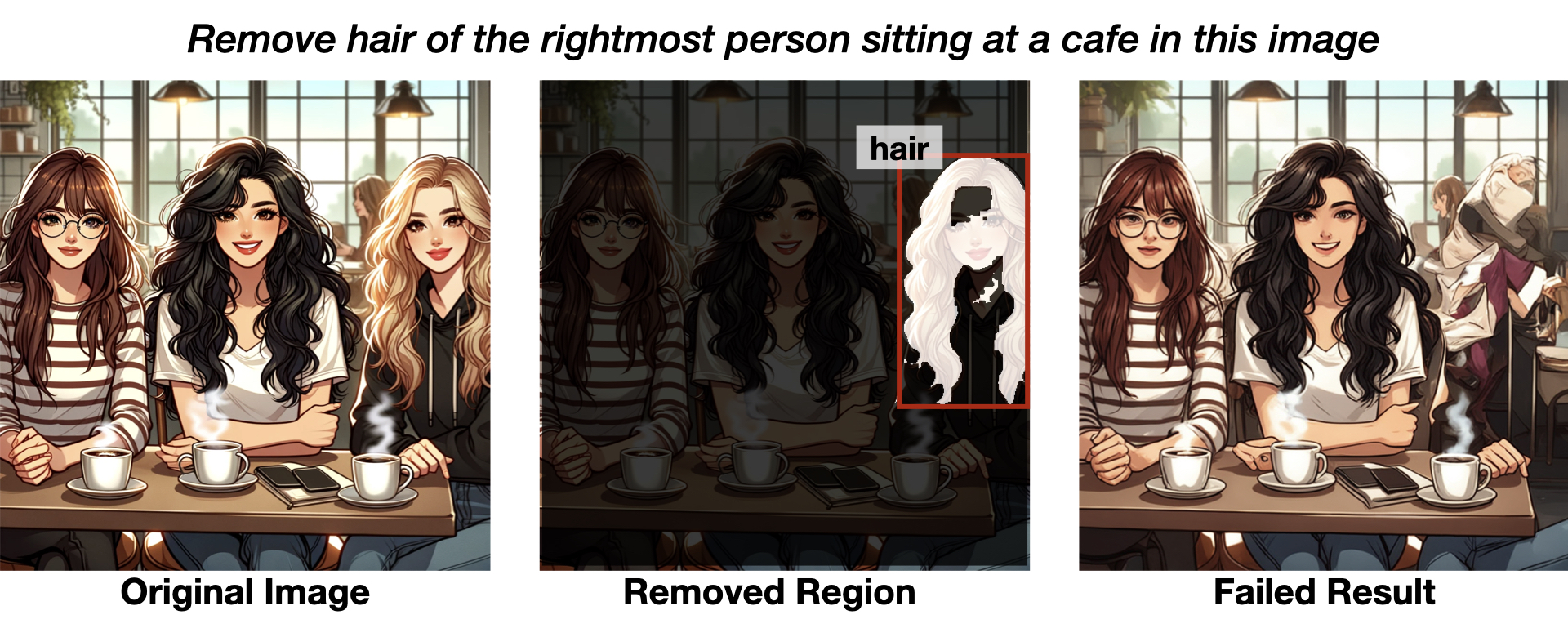}
\vspace{-16pt}
\caption{SLD struggles with objects of complex shapes, as the SAM module may unintentionally segment adjacent parts during the process.}
\vspace{-6pt}
\label{fig:failurecase}
\end{figure}
}

\def\figdetector#1{
\begin{figure}[#1]
\centering
\includegraphics[width=1.0\linewidth]{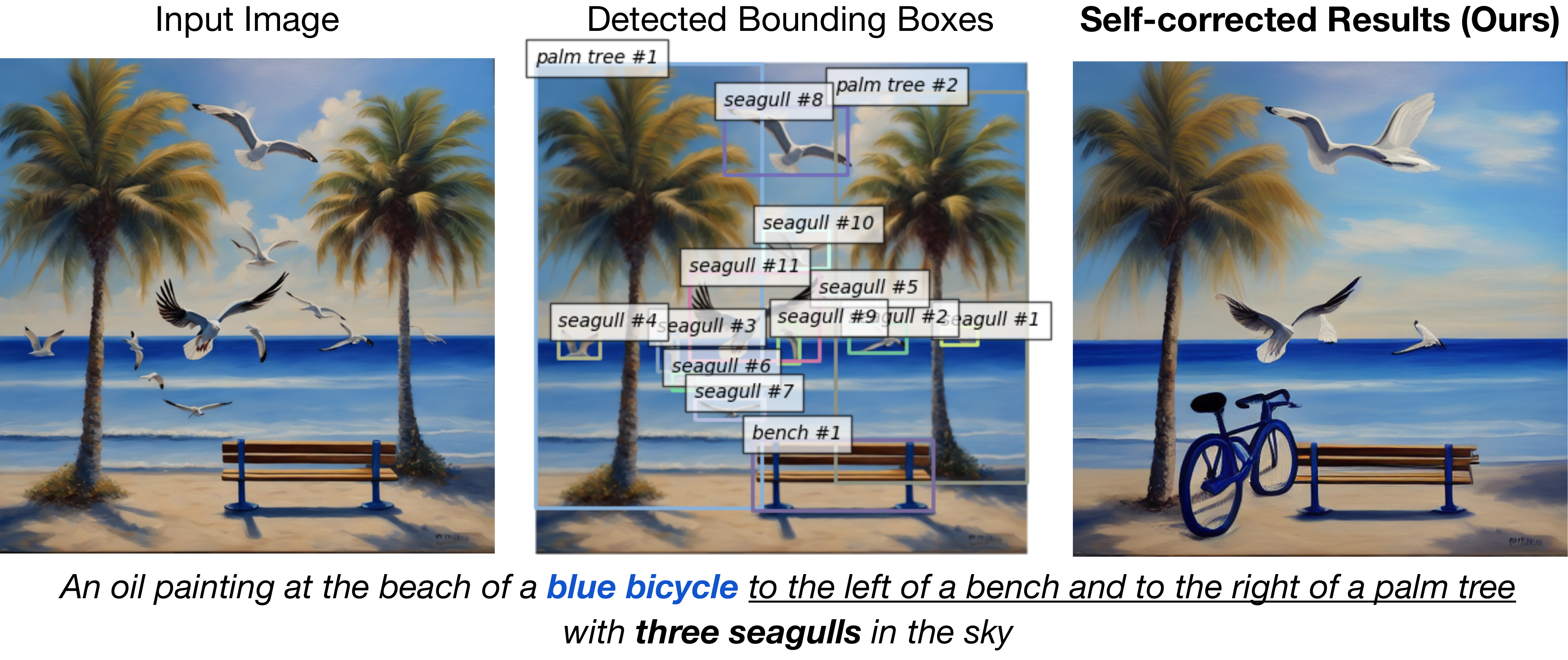}
\vspace{-16pt}
\caption{Leveraging the advanced localization abilities of OWL-ViT v2 open-vocabulary detectors, we accurately identify all seagulls in the image, enabling selective removal to align with the user's prompt.}
\vspace{-6pt}
\label{fig:detector_importance}
\end{figure}
}

\def\figedit#1{
\begin{figure}[#1]
\centering
\includegraphics[width=1.0\linewidth]{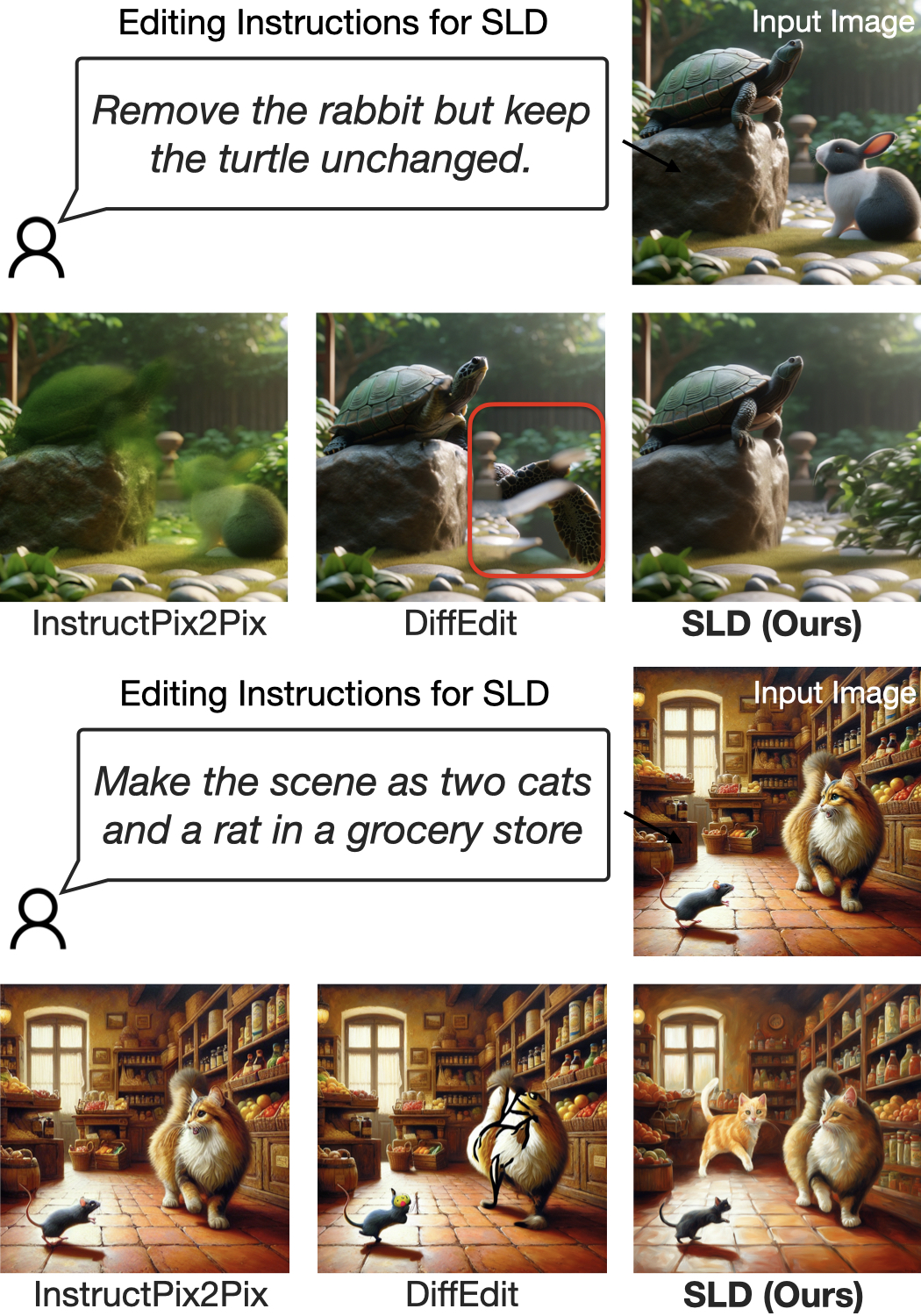}
\vspace{-16pt}
\caption{We demonstrate that current diffusion-based image editing methods, such as InstructPix2Pix and DiffEdit, face challenges in fundamental operations like object deletion (top example) and addition (bottom example). In contrast, our proposed SLD pipeline can easily handle these tasks. As highlighted in the red box, despite DiffEdit's ability to identify the object for removal, it falls short in generating a plausible background.}
\vspace{-6pt}
\label{fig:editing_supp}
\end{figure}
}

\def\tabMain#1{
\begin{table*}[#1]
\centering
\begin{tabular}{lccccc@{\hspace{3pt}}c}\toprule
& \multicolumn{6}{c}{Accuracy} \\
\cmidrule(lr){2-7}
Method & Negation & Numeracy & Attribute & Spatial & Average & \\\midrule
{\small MultiDiffusion \citep{bar2023multidiffusion}}
&\ \ 29\% & 28\% & 26\% & 39\% & 30.5\%\\
{\small Backward Guidance \citep{chen2023training}}
& \ \ 22\% & 37\% & 26\% & 67\% & 38.0\% \\
{\small BoxDiff \citep{xie2023boxdiff}}
& \ \ 22\% & 30\% & 37\% & 71\% & 40.0\% \\

{\small LayoutGPT + GLIGEN \citep{feng2023layoutgpt,li2023gligen}} 
& \ \ 36\% & 65\% & 26\% & 78\% & 51.3\% \\
\midrule
{\small DALL-E 3 \cite{dalle3}}
& \ \ 25\% & 38\% & 74\% & 71\% & 52.0\%\\
\rowcolor{LightGreen}{\small \quad + 1-round SLD (Ours)}
& \ \ 90\% & 61\% & \ \textbf{80\%} & 83\% & 78.5\% & {\small \textbf{\textcolor{Green}{(\textbf{+26.5})}}}\\
\midrule
{\small LMD+ \cite{lian2023llmgrounded}}
& \textbf{100\%} & 82\% & 49\% & 86\% & 79.3\%\\
\rowcolor{LightGreen}{\small \quad + 1-round SLD (Ours)}
& \textbf{100\%} & \ \textbf{98\%} & 63\% & \ \textbf{92\%} & \ \textbf{88.3\%} & {\small \textbf{\textcolor{Green}{(\textbf{+~~9.0})}}}\\
\bottomrule
\end{tabular}
\caption{Our method can be applied on various image generation methods and improves the generation accuracy by a large margin.\vspace{-6pt}}
\label{tab:main}
\end{table*}
}

\def\tabMultiRound#1{
\begin{table}[#1]
\centering
\setlength\tabcolsep{2pt}\resizebox{1.0\linewidth}{!}{
\begin{tabular}{lccccc@{\hspace{3pt}}l}\toprule
& \multicolumn{6}{c}{Accuracy} \\
\cmidrule(lr){2-7}
Method & \small{Negation} & \small{Numeracy} & \small{Attribute} & \small{Spatial} & \small{Average} & \\\midrule
{\small SD \citep{rombach2022high}}
& 19\% & 38\% & 24\% & 33\% & 28.5\%\\
\rowcolor{LightGreen}{\small \quad + 1-round SLD}
& 69\% & 55\% & 25\% & 69\% & 54.5\% & {\small \textbf{\textcolor{Green}{(\textbf{+26.0})}}}\\
\rowcolor{LightGreen}{\small \quad + 2-round SLD}
& \textbf{73\%} & \textbf{61\%} & \textbf{31\%} & \textbf{75\%} & \textbf{60.0\%} & {\small \textbf{\textcolor{Green}{(\textbf{+31.5})}}}\\
\bottomrule
\end{tabular}

}

\caption{While the majority of errors are typically rectified in the first round, multi-round correction consistently outperforms a single-round approach.\vspace{-6pt}}
\label{tab:multiround}
\end{table}
}

\def\tabsupp#1{
\begin{table*}[#1]
\centering
\begin{tabular}{lccccc@{\hspace{3pt}}l}\toprule
& \multicolumn{6}{c}{Accuracy} \\
\cmidrule(lr){2-7}
Method & Negation & Numeracy & Attribute & Spatial & Average & \\\midrule

{\small DALL-E 3 \cite{dalle3}}
& \ \ 25\% & 38\% & 74\% & 71\% & 52.0\%\\
\rowcolor{LightGreen}{\small \quad + 1-round SLD (OWL-ViT v1)}
& \ \ 50\% & 51\% & 71\% & 82\% & 63.5\% &{\small \textbf{\textcolor{Green}{(\textbf{+11.5})}}}\\
\rowcolor{LightGreen}{\small \quad + 1-round SLD (OWL-ViT v2)}
& \ \ 90\% & 61\% & \ \textbf{80\%} & 83\% & 78.5\% &{\small \textbf{\textcolor{Green}{(\textbf{+26.5})}}}\\
\midrule
{\small LMD+ \cite{lian2023llmgrounded}}
& \textbf{100\%} & 82\% & 49\% & 86\% & 79.3\%\\
\rowcolor{LightGreen}{\small \quad + 1-round SLD (OWL-ViT v1)}
& \textbf{100\%} & 85\% & 59\% & 89\% & 83.3\% &{\small \textbf{\textcolor{Green}{(\textbf{+~~4.0})}}}\\
\rowcolor{LightGreen}{\small \quad + 1-round SLD (OWL-ViT v2)}
& \textbf{100\%} & \ \textbf{98\%} & 63\% & \ \textbf{92\%} & \ \textbf{88.3\%} &{\small \textbf{\textcolor{Green}{(\textbf{+~~9.0})}}}\\
\bottomrule
\end{tabular}
\caption{Our method can be applied on various image generation methods and improves the generation accuracy by a large margin.\vspace{-6pt}}
\label{tab:supp_main}
\end{table*}
}

\footnotetext[1]{*Equal contribution.} \footnotetext[2]{$^\dagger$Equal advising.}
\begin{abstract}
Text-to-image generation has witnessed significant progress with the advent of diffusion models. Despite the ability to generate photorealistic images, current text-to-image diffusion models still often struggle to accurately interpret and follow complex input text prompts. In contrast to existing models that aim to generate images only with their best effort, we introduce Self-correcting LLM-controlled Diffusion (SLD). SLD is a framework that generates an image from the input prompt, assesses its alignment with the prompt, and performs self-corrections on the inaccuracies in the generated image. Steered by an LLM controller, SLD turns text-to-image generation into an iterative closed-loop process, ensuring correctness in the resulting image. SLD is not only training-free but can also be seamlessly integrated with diffusion models behind API access, such as DALL-E 3, to further boost the performance of state-of-the-art diffusion models. Experimental results show that our approach can rectify a majority of incorrect generations, particularly in generative numeracy, attribute binding, and spatial relationships. Furthermore, by simply adjusting the instructions to the LLM, SLD can perform image editing tasks, bridging the gap between text-to-image generation and image editing pipelines. We will make our code available for future research and applications.

\end{abstract}    

\section{Introduction}

Text-to-image generation has made remarkable advancements, especially with the advent of diffusion models. However, these models often struggle with interpreting complex input text prompts, particularly those that require skills such as understanding the concept of numeracy, spatial relationships, and attribute binding with multiple objects. Despite the astonishing scaling of model sizes and training data, these challenges, as illustrated in \cref{fig:teaser}, are still present in state-of-the-art open-source and proprietary diffusion models.

Several research and engineering efforts aim to overcome these limitations. For instance, methods such as DALL-E 3 \cite{dalle3} focus on the diffusion training process and incorporate high-quality captions into the training data at a massive scale. However, this approach not only incurs substantial costs but also frequently falls short in generating accurate images from complicated user prompts, as shown in \cref{fig:teaser}. Other work harnesses the power of external models for a better understanding of the prompt in the inference process before the actual image generation. For example, \cite{lian2023llmgrounded,feng2023layoutgpt} leverages Large Language Models (LLMs) to pre-process textual prompts into structured image layouts and thus ensures the preliminary design aligns with the user's directives. However, such integration does not resolve the inaccuracies produced by the downstream diffusion models, particularly in images with complex scenarios like multiple objects, cluttered arrangements, or detailed attributes.

Drawing inspiration from the from the process of a human painting and a diffusion model in generating images, we observe a key distinction in their approach to creation. Consider a human artist tasked with painting a scene featuring two cats. Throughout the painting process, the artist remains cognizant of this requirement, ensuring that two cats are indeed present before considering the work complete. Should the artist find only one cat depicted, an additional one would be added to meet the prompt's criteria. This contrasts sharply with current text-to-image diffusion models, which operate on an \textit{open-loop} basis. These models generate images through a predetermined number of diffusion steps and present the output to the user, regardless of its alignment with the initial user prompt. Such a process, irrespective of scaling training data or LLM pre-generation conditioning, lacks a robust mechanism to ensure the final image aligns with the user's expectations.

In light of this, we propose our method \textbf{S}elf-correcting \textbf{L}LM-controlled \textbf{D}iffusion (\textbf{SLD}) that performs \textit{self-checks} to confidently offer users guarantees of the alignment between the prompt and the generated images. Departing from conventional single-round generation methods, SLD is a novel \textit{closed-loop} approach that equips diffusion models with the ability to iteratively identify and rectify errors. Our SLD framework, illustrated in \cref{fig:main}, contains two main components: LLM-driven object detection as well as LLM-controlled assessment and correction.

The SLD pipeline follows a standard text-to-image generation setting.  Given a textual prompt that outlines the desired image, SLD begins with calling an image generation module (\eg, the aforementioned open-loop text-to-image diffusion models) to generate an image in a best-effort fashion. Given that these open-loop generators do not guarantee an output that aligns perfectly with the prompt, SLD then conducts a thorough evaluation of the produced image against the prompt, with an LLM parsing key phrases for an open-vocabulary detector to check. 
Subsequently, an LLM controller takes the detected bounding boxes and the initial prompt as input, checks for potential mismatches between the detection results and the prompt requirements, suggesting appropriate self-correction operations, such as adding, moving, and removing objects. Finally, utilizing a base diffusion model (\eg, Stable Diffusion \cite{rombach2022high}), SLD employs latent space composition to implement these adjustments, thereby ensuring that the final image accurately reflects the user's initial text prompt.

Notably, our pipeline does not pose restrictions on the source of the initial generation and thus is applicable to images generated by proprietary models, such as DALL-E 3 \cite{dalle3}, through APIs, as shown in \cref{fig:teaser}. Furthermore, none of the self-correction operations require any additional training on our base diffusion model \cite{rombach2022high}, which easily allows our method to be applied to various diffusion models without the costs of external human annotation or training.

We demonstrate that our SLD framework can achieve significant improvement over current diffusion-based methods on cpmplex prompts with the LMD benchmark \cite{lian2023llmgrounded}. The results show that our method is able to surpass LMD+, which is a strong baseline that already leverages LLM in the image process generation, by $9.0\%$. More importantly, with DALL-E 3 for initial generation, the generated images from our method achieve $26.5\%$ performance gains compared to ones before self-correction.

Finally, since the SLD pipeline is agnostic to the initially generated image, it can easily be transformed into an image editing pipeline by simply changing the prompts to the LLM. While text-to-image generation and image editing are often treated as distinct tasks by the generative modeling community, our SLD is able to perform these two tasks with a unified pipeline. \textbf{We list our key contributions below:}

\begin{enumerate}
 \item SLD is the first to integrate a detector and an LLM to \textit{self-correct} generative models, ensuring accurate generation without extra training or external data.
 \item SLD offers a unified solution for both image generation and editing, enabling enhanced text-to-image alignment for any image generator (\eg, DALL-E 3) and object-level editing on any images.
 \item Experimental results show that our approach can correct a majority of incorrect generations, particularly in aspects of numeracy, attribute binding, and spatial relationships.
\end{enumerate}
\noindent We will release our code for future research and applications.

\figMain{t!}

\section{Related Work}

\subsection{Text-to-Image Diffusion Models}

Diffusion-based text-to-image generation has advanced significantly. Initial studies \cite{rombach2022high, ramesh2022hierarchical, saharia2022photorealistic} showed diffusion models' ability to create high-quality images, but they struggle with complex prompts. Subsequent research \cite{zhang2023adding,li2023gligen,yang2023reco,chen2023training,yang2023paint,patashnik2023localizing} has incorporated additional inputs such as keypoints and bounding boxes to control the diffusion generation process.

Recent advancements have incorporated LLMs to control the generation of diffusion models, bypassing the need for additional complementary information as inputs \cite{lian2023llmgrounded,lian2023llm,feng2023layoutgpt,lin2023videodirectorgpt,zhang2023controllable}. In these approaches, LLMs play a central role in directly interpreting user textual prompts and managing the initial layout configuration. Despite some progress, these models often operate in an open-loop fashion, producing images in one iteration that cannot guarantee the generated images align with user prompts.

Unlike prior work, SLD is the first closed-loop diffusion-based generation method. Integrating advanced object detectors and LLMs, SLD performs iterative self-checking and correction, significantly enhancing text-to-image alignment. This improvement spans numeracy, attribute binding to multiple objects, and spatial reasoning, applicable to various models, including those like DALL-E 3.

\subsection{Diffusion-based Image Editing}

Recent advancements in text-to-image diffusion models have significantly expanded their applications in image editing, encompassing both global and local editing tasks. Techniques like Prompt-2-prompt \cite{hertz2022prompt} and InstructPix2Pix \cite{brooks2023instructpix2pix} specialize in global edits, such as style transformations. Conversely, methods like SDEdit \cite{meng2022sdedit}, DiffEdit \cite{couairon2023diffedit}, and Plug-and-Play \cite{tumanyan2023plug} focus on local edits, targeting specific areas within images. Despite their progress, these methods often struggle with precise object-level manipulation and tasks that require spatial reasoning, such as resizing or repositioning objects. While recent approaches like Self-Guidance \cite{epstein2023selfguidance} offer fine-grained operations, they still necessitate user inputs for specific coordinates when moving or repositioning objects.

Unlike these methods only focusing on diffusion models, SLD introduces the combination of detectors and LLMs in the loop of editng, enabling fine-grained editing with only user prompts. Also, SLD excels in a variety of object-level editing tasks, including adding, replacing, moving, and modifying attributes, swapping, and so on, demonstrating a notable improvement in both ease of use and editing capabilities.

\section{Self-correcting LLM-controlled Diffusion}

In this section, we introduce our \textbf{S}elf-correcting \textbf{L}LM-controlled \textbf{D}iffusion (\textbf{SLD}) framework. SLD consists of two main components: LLM-driven object detection~(\cref{ssec:object_detection}) as well as LLM-controlled assessment and correction~(\cref{ssec:llm_controlled_image_analysis}).
Moreover, with a simple change of the LLM instructions, we show that SLD is applicable to image editing, unifying text-to-image generation and editing as discussed in \cref{ssec:unified_generation}. The complete pipeline is shown in \cref{alg:self-correct-image-generation}.

\figPrompt{t!}

\subsection{LLM-driven Object Detection}
\label{ssec:object_detection}

Our SLD framework starts with LLM-driven object detection, which extracts information necessary for downstream assessment and correction. As shown with green arrows in \cref{fig:main}, the LLM-driven object detection includes two steps: \textbf{1)} We leverage an LLM as a parser that parses the user prompt and outputs key phrases that are potentially relevant to image assessment. \textbf{2)} The phrases are then passed into an open-vocabulary object detector. The detected boxes are supposed to contain information that supports the assessment of whether the image is aligned with the specifications in the user prompt.

In the initial step, an LLM parser is directed to extract a list of key object details, denoted as $S$, from the user-provided text prompt $P$. This parser, aided by text instructions and in-context examples, can easily accomplish this as shown in \cref{fig:prompt} (a). For a user prompt that includes phrases like ``a green motorcycle" and ``a blue motorcycle," the LLM is expected to identify and output ``green" and ``blue" as attributes associated with ``motorcycle." When the prompt references objects without specific quantities or attributes, such as ``a monkey" and ``a raccoon," these descriptors are appropriately left blank. Importantly, the LLM's role is not limited to merely identifying object nouns; it also entails identifying any associated quantities or attributes.

In the second step, an open-vocabulary detector processes the list of key object information, $S$, parsed in the first step, to detect and localize objects within the image. We prompt the open-vocabulary object detector with queries formatted as \texttt{image of a/an [attribute] [object name]}, where the ``attribute'' and ``object name'' are sourced from the parser's output. The resulting bounding boxes, $B_{curr}$, are then organized into a list format like \texttt{[("[attribute] [object name] [\#object ID]", [x, y, w, h])]} for further processing. A special case is when the prompt poses constraints on the object quantity. For cases where attributed objects (e.g., ``blue dog'') fall short compared to the required quantities, a supplementary count of non-attributed objects (e.g., ``dog'') is provided to provide context for the subsequent LLM controller deciding whether to add more ``blue dogs" or simply alter the color of existing dogs to blue. We will explain these operations, including object addition and attribute modification, in greater detail in \cref{ssec:latent_ops}.

\subsection{LLM-controlled Analysis and Correction}
\label{ssec:llm_controlled_image_analysis}

We use an LLM controller for image analysis and the subsequent correction. The controller, given the user prompt $P$ and detected boxes $B_{curr}$, is asked to analyze whether the image, represented by objects bounding boxes, aligns with the description of the user prompt and offer a list of corrected bounding boxes $B_{next}$, as shown in \cref{fig:prompt} (b). 

SLD then programmatically analyzes the inconsistencies between the refined and original bounding boxes to output a set of editing operations $Ops$, which includes addition, deletion, repositioning, and attribute modification. However, a simple set-of-boxes representation does not carry correspondence information, which does not allow an easy way to compare the input and the output layout of the LLM controller when multiple boxes share the same object name. For example, when there are two cat boxes in both the model input and the model output, whether one cat box corresponds to which cat box in the output layout is unclear. Rather than introducing another algorithm to guess the correspondence, we propose to let the LLM output correspondence with a very simple edit: we give an object ID to each bounding box, with the number increasing within each object type, as a suffix added after the object name. In the in-context examples, we demonstrate to the LLM that the object should have the same name and object ID before and after the proposed correction.

\subsubsection{Latent Operations for Training-Free Image Correction}
\label{ssec:latent_ops}
The LLM controller outputs a list of correction operations to apply. For each operation, we first transform the original image into latent features. Our approach then executes a series of operations $Ops$, such as addition, deletion, repositioning, and attribute modification, applied to these latent layers. We explain how each operation is performed below. %

\figLatent{h!}

\Algo{t!}

\noindent \textbf{Addition.} Inspired by \cite{lian2023llmgrounded}, the addition process entails two phases: pre-generating an object and integrating its latent representation into the original image's latent space. Initially, we use a diffusion model to create an object within a designated bounding box, followed by precise segmentation using models (\eg, SAM \cite{kirillov2023segment}). This object is then processed through a backward diffusion sequence with our base diffusion model, yielding masked latent layers corresponding to the object, which are later merged with the original canvas.

\noindent \textbf{Deletion} operation begins with SAM refining the boundary of the object within its bounding box. The latent layers associated with these specified regions are then removed and reset with Gaussian noise. This necessitates a complete regeneration of these areas during the following forward diffusion process.

\noindent \textbf{Repositioning} involves modifying the original image to align objects with new bounding boxes, taking care to preserve their original aspect ratios. The initial steps include shifting and resizing the bounding box in the image space. Following this, SAM refines the object boundary, succeeded by a backward diffusion process to generate its relevant latent layers, similar to the approach in the addition operation. Latent layers corresponding to the excised parts are replaced with Gaussian noise, while the newly added sections are integrated into the final image composition. An important consideration in repositioning is conducting object resizing in the image space rather than the latent space to maintain high-quality results.

\noindent \textbf{Attribute modification} starts with SAM refining the object boundary within the bounding box, followed by applying attribute modifications such as DiffEdit \cite{couairon2023diffedit}. The base diffusion model then reverses the image, producing a series of masked latent layers ready for final composition.

After editing operations on each object, we proceed to the recomposition phase as shown in \cref{fig:latent}. In this phase, while latents for removed or repositioned regions are reinitialized with Gaussian noise, the latents for added or modified latents are updated accordingly. For regions with multiple overlapping objects, we place the larger masks first to ensure the visibility of the smaller objects.

The stack of modified latent then undergoes a final forward diffusion process, which begins with steps in which regions not reinitialized with Gaussian noise are frozen (\ie, forced to align with the unmodified latent at the same step). This is crucial for the accurate formation of updated objects while maintaining background consistency at the same time. The procedure finishes with several steps where everything is allowed to change, resulting in a visually coherent and correct image.

\tabMain{t!}
\subsubsection{Termination of the Self-Correction Process}

Even though we observe that one round of generation is often enough for a majority of the cases that we encountered, subsequent rounds could still benefit the performance in terms of correctness further, making our self-correction an iterative process.

Determining the optimal number of self-correction rounds is critical for balancing efficiency and accuracy. As outlined in \cref{alg:self-correct-image-generation}, our method sets a maximum number of attempts on the correction rounds to ensure the process finishes within a reasonable amount of time.

The process completes when the LLM outputs the same layout as the input (\ie, if the bounding boxes suggested by the LLM controller ($B_{next}$) align with the current detected bounding boxes ($B_{curr}$)), or when the maximum rounds of generation are reached, which indicates that the method is unable to make a correct generation for the prompt. This iterative process provides guarantees on the correctness of the image, up to the accuracy of the detector and the LLM controller, ensuring it aligns closely with the initial text prompt. We explore the efficacy of multi-round corrections in \cref{ssec:discussion}.

\figVis{t!}

\subsection{Unified text-to-image generation and editing}
\label{ssec:unified_generation}

In addition to self-correcting image generation models, our SLD framework is readily adaptable for image editing applications, requiring only minimal modifications. A key distinction is in the format of user input prompts. Unlike image generation, where users provide scene descriptions, image editing requires users to detail both the original image and the desired changes. For instance, to edit an image with two apples and a banana by replacing the banana with an orange, the prompt could be: ``Replace the banana with an orange, while keeping the two apples unchanged."

The editing process is similar to our self-correction mechanism. The LLM parser extracts key objects from the user's prompt. These objects are then identified by the open-vocabulary detector, establishing a list of current bounding boxes. The editing-focused LLM controller, equipped with specific task objectives, guidelines, and in-context examples, analyzes these inputs. It proposes updated bounding boxes and corresponding latent space operations for precise image manipulation.

SLD's ability to perform detailed, object-level editing distinguishes it from existing diffusion-based methods like InstructPix2Pix \cite{brooks2023instructpix2pix} and prompt2prompt \cite{hertz2022prompt}, which mainly address global image style changes. Also, SLD outperforms tools like DiffEdit \cite{couairon2023diffedit} and SDEdit \cite{meng2022sdedit}, which are restricted to object replacement or attribute adjustment, by enabling comprehensive object repositioning, addition, and deletion with exact control. Our comparative analysis in \cref{ssec:editing_application} will further highlight SLD's superior editing capabilities over existing methods.

\section{Experiments}
\subsection{Comparison with Image Generation Methods}

\noindent\textbf{Setup.} We evaluate the performance of the SLD framework with the LMD benchmark~\cite{lian2023llmgrounded}, which is specifically designed to evaluate generation methods on complex tasks such as handling negation, numeracy, accurate attribute binding to multiple objects, and spatial reasoning. For each task, 100 programmatically generated prompts are fed into various text-to-image generation methods to produce corresponding images. We evaluate the images generated by our method and the baselines with open-vocabulary detector OWL-ViT v2 \cite{minderer2023scaling} for a robust quantitative evaluation of the alignment between the input prompts and the generated images. We compared SLD with several leading text-to-image diffusion methods, such as Multidiffusion \cite{bar2023multidiffusion}, BoxDiff \cite{xie2023boxdiff}, LayoutGPT \cite{feng2023layoutgpt}, LMD+ \cite{lian2023llm}, and DALL-E 3 \cite{dalle3}. To ensure fair comparisons, all models incorporating LLMs used the same GPT-4 model. For our SLD implementation, we utilized LMD+ as the base model for latent space operations and OWL-ViT v2 for the open-vocabulary object detector.

\noindent\textbf{Results.} As shown in \cref{tab:main}, applying the SLD method to both open-source (LMD+) and proprietary models (DALL-E 3) significantly enhances their performance in terms of generation correctness. \textbf{For negation tasks,} as LMD+ converts user prompts containing ``without'' information into negative prompts, which already achieves a remarkable 100\% accuracy without SLD integration. 
In contrast, even though DALL-E 3 also uses an LLM to rewrite the prompt, it still fails for some negation cases, likely because the LLM simply puts the negation keyword (\eg, ``without'') into the rewritten prompt. In this case, our SLD method can automatically rectify most of these errors. %
\textbf{For numeracy tasks,} integrating SLD with LMD+ results in a significant improvement, with up to 98\% accuracy. We noted that DALL-E 3 often struggles to generate an image with the correct number of objects. However, this issue is substantially mitigated by SLD, which enhances performance by over 20\%. \textbf{For attribute binding tasks,} SLD improves the performance of both DALL-E 3 and LMD+ by 6\% and 14\%, respectively. Notably, DALL-E 3 initially outperforms LMD+ in this task, likely due to its training on high-quality image caption datasets. Finally, \textbf{for spatial reasoning tasks,} the integration of SLD with both LMD+ and DALL-E 3 demonstrates enhanced performance by 12\% and 6\%, respectively. %

\figEditCompare{t!}
\figEdit{t!}
\figFailure{h}

\subsection{Application to Image Editing}
\label{ssec:editing_application}

As discussed in  \cref{ssec:unified_generation}, SLD excels in fine-grained image editing over existing methods. As demonstrated in \cref{fig:edit_comparison}, our integration of an open-language detector with LLMs enables precise modifications within localized latent space regions. SLD adeptly performs specific edits, like seamlessly replacing an apple with a pumpkin, while preserving the integrity of surrounding objects. In contrast, methods like InstructPix2Pix \cite{brooks2023instructpix2pix} are confined to global transformations, and DiffEdit \cite{couairon2023diffedit} often fails to accurately locate objects for modification, leading to undesired results.

Furthermore, as exemplified in \cref{fig:edit}, SLD supports a wide array of editing instructions, including counting control (such as adding, deleting, or replacing objects), attribute modification (like altering colors or materials), and intricate location control (encompassing object swapping, resizing, and moving). A standout example is featured in the ``Object Resize'' column of the first row, where SLD precisely enlarges the cup on the table by an exact factor of 1.25$\times$. We encourage readers to verify this with a ruler for a clear demonstration of our method's precision. This level of precision stems from the detector's exact object localization coupled with the LLMs' ability in reasoning and suggestions for new placements. Such detailed control over spatial adjustments is unmatched by any previous method, highlighting SLD's contributions to fine-grained image editing.

\subsection{Discussion}
\label{ssec:discussion}

\noindent \textbf{Multi-round self-correction.} Our analysis in \cref{tab:multiround} highlights the benefits of multi-round self-correction and the fact that the first round correction is always the most effective one and has a marginal effect. The first round of corrections substantially mitigates issues inherent in Stable Diffusion \cite{rombach2022high}. Then, a second round of correction still yields significant improvements across all four tasks.

\tabMultiRound{t!}

\noindent \textbf{Limitations and future work.} A limitation of our method is illustrated in \cref{fig:failurecase}, where SLD fails to accurately remove a person's hair. In this instance, despite the successful identification and localization of the hair, the complex nature of its shape poses a challenge to the SAM module used for region selection, resulting in the unintended removal of the person's face in addition to the hair. However, since the person's cloth is not removed, the base diffusion model fails to generate a natural composition. This suggests that a better region selection method is needed for further improvements in the generation and editing quality.

\section{Conclusion}

We introduce the Self-correcting Language-Driven (SLD) framework, a pioneering self-correction system using detectors and LLMs to significantly enhance text-to-image alignment. This method not only sets a new SOTA in image generation benchmark but is also compatible with various generative models, including DALL-E 3. Also, SLD extends its utility to image editing applications, offering fine-grained object-level manipulation that surpasses existing methods.
{
    \small
    \bibliographystyle{ieeenat_fullname}
    \bibliography{main}
}

\clearpage
\appendix

\tabsupp{t!}

\section{More Comparisons with Image Editing Methods}

In \cref{fig:edit_comparison} of the main paper, we showcased an example of object replacement contrasting our SLD method with previous approaches. In this section, we provide more visual examples in \cref{fig:editing_supp} that highlight the differences between SLD and prior diffusion-based image editing methods. While InstructPix2Pix~\cite{brooks2023instructpix2pix} is confined to pixel-aligned changes (\eg, style changes), DiffEdit~\cite{couairon2023diffedit} often fails with precise object-level edits. Our SLD framework excels in these detailed, fine-grained editing tasks.

\section{Comparison between LMMs and Detector-LLM Combinations}

In our main paper, we introduce a self-correction framework that utilizes both LLMs and object detectors for image assessment, followed by the provision of correction suggestions. With the rapid evolution of Large Multimodal Models~(LMMs) such as GPT-4V \cite{gpt4v} and LLaVA \cite{liu2023llava,liu2023improvedllava}, we have investigated the feasibility of using an LMM to conduct image assessment. In this setup, the LMM evaluates images generated by the open-loop generator alongside user prompts, aiming to provide precise, object-level editing recommendations.

However, GPT-4V's analysis of the ``princess and dwarfs" image from DALL-E 3 (refer to \cref{fig:teaser} in our paper) reveals inaccuracies, as the model miscounts the characters, identifying seven dwarfs rather than the actual five, and struggles to define precise bounding box coordinates. In contrast, as highlighted in \cref{fig:detector_importance}, the latest generation open-vocabulary detector, OWL-ViT v2, demonstrates remarkable proficiency in detecting minute objects (seagulls in the sky), which is vital for the correction process. These limitation underscores our current approach, which combines detectors with LLMs for more accurate assessment and editing suggestions. Nevertheless, the potential of integrating advanced LMMs for streamlined image generation and editing remains a compelling and promising direction for future research and development.

\figedit{t}
\figdetector{t}

\section{Comprehensive Image Generation Results}

In the main paper, we demonstrate the significant performance gain achieved by integrating the OWL-ViT v2 open-vocabulary detector into our SLD pipeline. Since this detector is also used in the benchmark, we also explore using other detectors in our setting, which makes our self-correction detector distinct from the detector used in the evaluation benchmark.

\cref{tab:supp_main} presents a comparison between the results obtained using OWL-ViT v1 \cite{minderer2022simple} and OWL-ViT v2 \cite{minderer2023scaling} detectors within our framework. The results indicate that SLD consistently enhances overall accuracy compared to the baseline text-to-image generators, irrespective of the detector used. The marginal reduction in performance gains when substituting the v2 detector can be attributed to two main factors: \textbf{1)} the relatively inferior detection capabilities of the alternate detector employed in the SLD, and \textbf{2)} the variations in object recognition between the two detectors. This situation parallels real-world experiences, where individual perceptions and recognition of objects or attributes can vary significantly. For instance, a bowl perceived as distinctly blue by one might be seen as less blue by another. These perceptual variances contribute to the marginally lower attribute binding scores of SLD compared to original DALL-E 3 results. Despite these discrepancies, the overall accuracy of our method, especially in areas such as numeracy and spatial reasoning, confirms the effectiveness of SLD, \textit{regardless of the specific detector used in our self-correction pipeline}.

\section{Our Prompts and Instructions to the LLM}

As outlined in the main paper,  we leverage two LLMs to steer the self-correction process: we employ one LLM parser to identify key objects from user prompts and another LLM controller to propose bounding box adjustments. The specific prompts for both the parser and the controller are detailed in \cref{tab:parser_prompt} and \cref{tab:controller_prompt}, respectively. We also provide in-context examples for the LLM controller, tailored for self-correcting generation and image editing, in \cref{tab:gen_example} and \cref{tab:edit_example}, respectively.

In crafting our LLM prompts, we emphasized clarity in defining the roles and guidelines for the LLM, drawing inspiration from prior work \cite{white2023prompt}. Notably, we discovered that GPT-4 possesses the capability to manipulate bounding box coordinates, a task that involves mathematical reasoning. We achieved improved results by guiding the model to employ chain-of-thought reasoning \cite{wei2022chain}, where the model explicates its reasoning process during generation. This approach yielded more accurate suggestions compared to instances where the model's reasoning was not explicitly stated.

\begin{table*}[h!]
\setlength\tabcolsep{0pt}
\centering
\begin{tabular*}{\linewidth}{@{\extracolsep{\fill}} l }\toprule
\begin{lstlisting}[style=myverbatim]
# Your Role: Excellent Parser

## Objective: Analyze scene descriptions to identify objects and their attributes.

## Process Steps
1. Read the user prompt (scene description).
2. Identify all objects mentioned with quantities.
3. Extract attributes of each object (color, size, material, etc.).
4. If the description mentions objects that shouldn't be in the image, take note at the negation part.
5. Explain your understanding (reasoning) and then format your result (answer / negation) as shown in the examples.
6. Importance of Extracting Attributes: Attributes provide specific details about the objects. This helps differentiate between similar objects and gives a clearer understanding of the scene.

## Examples

- Example 1
    User prompt: A brown horse is beneath a black dog. Another orange cat is beneath a brown horse.
    Reasoning: The description talks about three objects: a brown horse, a black dog, and an orange cat. We report the color attribute thoroughly. No specified negation terms.
    Objects: [('horse', ['brown']), ('dog', ['black']), ('cat', ['orange'])]
    Negation: 

- Example 2
    User prompt: There's a white car and a yellow airplane in a garage. They're in front of two dogs and behind a cat. The car is small. Another yellow car is outside the garage.
    Reasoning: The scene has two cars, one airplane, two dogs, and a cat. The car and airplane have colors. The first car also has a size. No specified negation terms.
    Objects: [('car', ['white and small', 'yellow']), ('airplane', ['yellow']), ('dog', [None, None]), ('cat', [None])]
    Negation: 

- Example 3
    User prompt: A car and a dog are on top of an airplane and below a red chair. There's another dog sitting on the mentioned chair.
    Reasoning: Four objects are described: one car, airplane, two dog, and a chair. The chair is red color. No specified negation terms.
    Objects: [('car', [None]), ('airplane', [None]), ('dog', [None, None]), ('chair', ['red'])]
    Negation: 

- Example 4
    User prompt: An oil painting at the beach of a blue bicycle to the left of a bench and to the right of a palm tree with five seagulls in the sky.
    Reasoning: Here, there are five seagulls, one blue bicycle, one palm tree, and one bench. No specified negation terms.
    Objects: [('bicycle', ['blue']), ('palm tree', [None]), ('seagull', [None, None, None, None, None]), ('bench', [None])]
    Negation: 

- Example 5
    User prompt: A realistic photo of a scene without backpacks.
    Reasoning: The description clearly states no backpacks, so this must be acknowledged. The user provides the negative prompt of backpacks.
    Objects: [('backpacks', [None])]
    Negation: backpacks

Your Current Task: Follow the steps closely and accurately identify objects based on the given prompt. Ensure adherence to the above output format.

User prompt: {the input user prompt}
Reasoning: 
\end{lstlisting} \\\bottomrule
\end{tabular*}
\caption{Our full prompt for the LLM parser.}
\label{tab:parser_prompt}
\end{table*}

{
\begin{table*}[h!]
\setlength\tabcolsep{0pt}
\centering
\begin{tabular*}{\linewidth}{@{\extracolsep{\fill}} l }\toprule
\begin{lstlisting}[style=myverbatim]
# Your Role: Expert Bounding Box Adjuster

## Objective: Manipulate bounding boxes in square images according to the user prompt while maintaining visual accuracy.

## Bounding Box Specifications and Manipulations
1. Image Coordinates: Define square images with top-left at [0, 0] and bottom-right at [1, 1].
2. Box Format: [Top-left x, Top-left y, Width, Height]
3. Operations: Include addition, deletion, repositioning, and attribute modification.

## Key Guidelines
1. Alignment: Follow the user's prompt, keeping the specified object count and attributes.
2. Boundary Adherence: Keep bounding box coordinates within [0, 1].
3. Minimal Modifications: Change bounding boxes only if they don't match the user's prompt.
4. Overlap Reduction: Minimize intersections in new boxes and remove the smallest, least overlapping objects.

## Process Steps
1. Interpret prompts: Read and understand the user's prompt.
2. Implement Changes: Review and adjust current bounding boxes to meet user specifications.
3. Explain Adjustments: Justify the reasons behind each alteration and ensure every adjustment abides by the key guidelines.
4. Output the Result: Present the reasoning first, followed by the updated objects section, which should include a list of bounding boxes in Python format.

## Examples

{In-context examples for self-correcting image generation or image editing}

Your Task: Carefully follow the provided guidelines and steps to adjust bounding boxes in accordance with the user's prompt. Ensure adherence to the above output format.

User prompt: {the input user prompt}
Current Objects: {a list of detected key objects}
Reasoning:
\end{lstlisting} \\\bottomrule
\end{tabular*}
\caption{Our full prompt for the LLM controller.}
\label{tab:controller_prompt}
\end{table*}
}

{
\begin{table*}[h!]
\setlength\tabcolsep{0pt}
\centering
\begin{tabular*}{\linewidth}{@{\extracolsep{\fill}} l }\toprule
\begin{lstlisting}[style=myverbatim]
# Examples

- Example 1
    User prompt: A realistic image of landscape scene depicting a green car parking on the left of a blue truck, with a red air balloon and a bird in the sky
    Current Objects: [('green car #1', [0.027, 0.365, 0.275, 0.207]), ('blue truck #1', [0.350, 0.368, 0.272, 0.208]), ('red air balloon #1', [0.086, 0.010, 0.189, 0.176])]
    Reasoning: To add a bird in the sky as per the prompt, ensuring all coordinates and dimensions remain within [0, 1].
    Updated Objects: [('green car #1', [0.027, 0.365, 0.275, 0.207]), ('blue truck #1', [0.350, 0.369, 0.272, 0.208]), ('red air balloon #1', [0.086, 0.010, 0.189, 0.176]), ('bird #1', [0.385, 0.054, 0.186, 0.130])]

- Example 2
    User prompt: A realistic image of landscape scene depicting a green car parking on the right of a blue truck, with a red air balloon and a bird in the sky
    Current Output Objects: [('green car #1', [0.027, 0.365, 0.275, 0.207]), ('blue truck #1', [0.350, 0.369, 0.272, 0.208]), ('red air balloon #1', [0.086, 0.010, 0.189, 0.176])]
    Reasoning: The relative positions of the green car and blue truck do not match the prompt. Swap positions of the green car and blue truck to match the prompt, while keeping all coordinates and dimensions within [0, 1].
    Updated Objects:  [('green car #1', [0.350, 0.369, 0.275, 0.207]), ('blue truck #1', [0.027, 0.365, 0.272, 0.208]), ('red air balloon #1', [0.086, 0.010, 0.189, 0.176]), ('bird #1', [0.485, 0.054, 0.186, 0.130])]

- Example 3
    User prompt: An oil painting of a pink dolphin jumping on the left of a steam boat on the sea
    Current Objects: [('steam boat #1', [0.302, 0.293, 0.335, 0.194]), ('pink dolphin #1', [0.027, 0.324, 0.246, 0.160]), ('blue dolphin #1', [0.158, 0.454, 0.376, 0.290])]
    Reasoning: The prompt mentions only one dolphin, but two are present. Thus, remove one dolphin to match the prompt, ensuring all coordinates and dimensions stay within [0, 1].
    Updated Objects: [('steam boat #1', [0.302, 0.293, 0.335, 0.194]), ('pink dolphin #1', [0.027, 0.324, 0.246, 0.160])]

- Example 4
    User prompt: An oil painting of a pink dolphin jumping on the left of a steam boat on the sea
    Current Objects: [('steam boat #1', [0.302, 0.293, 0.335, 0.194]), ('dolphin #1', [0.027, 0.324, 0.246, 0.160])]
    Reasoning: The prompt specifies a pink dolphin, but there's only a generic one. The attribute needs to be changed.
    Updated Objects: [('steam boat #1', [0.302, 0.293, 0.335, 0.194]), ('pink dolphin #1', [0.027, 0.324, 0.246, 0.160])]

- Example 5
    User prompt: A realistic photo of a scene with a brown bowl on the right and a gray dog on the left
    Current Objects: [('gray dog #1', [0.186, 0.592, 0.449, 0.408]), ('brown bowl #1', [0.376, 0.194, 0.624, 0.502])]
    Reasoning: The leftmost coordinate (0.186) of the gray dog's bounding box is positioned to the left of the leftmost coordinate (0.376) of the brown bowl, while the rightmost coordinate (0.186 + 0.449) of the bounding box has not extended beyond the rightmost coordinate of the bowl. Thus, the image aligns with the user's prompt, requiring no further modifications.
    Updated Objects: [('gray dog #1', [0.186, 0.592, 0.449, 0.408]), ('brown bowl #1', [0.376, 0.194, 0.624, 0.502])]
\end{lstlisting} \\\bottomrule
\end{tabular*}
\caption{Self-correction in-context examples for the LLM controller.}
\label{tab:gen_example}
\end{table*}
}

{
\begin{table*}[h!]
\setlength\tabcolsep{0pt}
\centering
\begin{tabular*}{\linewidth}{@{\extracolsep{\fill}} l }\toprule
\begin{lstlisting}[style=myverbatim]
## Examples

- Example 1
    User prompt: Move the green car to the right and make the blue truck larger in the image.
    Current Objects: [('green car #1', [0.027, 0.365, 0.275, 0.207]), ('blue truck #1', [0.350, 0.368, 0.272, 0.208])]
    Reasoning: To move the green car rightward, its x-coordinate needs to be increased from 0.027. The dimensions (height and width) of the blue truck must be enlarged. While adjusting bounding boxes, ensure they do not overlap excessively. All other elements remain unchanged.
    Updated Objects: [('green car #1', [0.327, 0.365, 0.275, 0.207]), ('blue truck #1', [0.350, 0.369, 0.472, 0.408])]

- Example 2
    User prompt: Swap the positions of a green car and a blue truck in this landscape scene with an air balloon.
    Current Output Objects: [('green car #1', [0.350, 0.369, 0.275, 0.207]), ('blue truck #1', [0.027, 0.365, 0.272, 0.208]), ('red air balloon #1', [0.086, 0.010, 0.189, 0.176])]
    Reasoning: Exchange locations of the car and truck to align the bottom right part; other objects remain unchanged.
    Updated Objects:  [('green car #1', [0.027, 0.365, 0.275, 0.207]), ('blue truck #1', [0.350, 0.364, 0.272, 0.208]), ('red air balloon #1', [0.086, 0.010, 0.189, 0.176])]

- Example 3
    User prompt: Change the color of the dolphin from blue to pink in this oil painting of a dolphin and a steamboat.
    Current Objects: [('steam boat #1', [0.302, 0.293, 0.335, 0.194]), ('blue dolphin #1', [0.027, 0.324, 0.246, 0.160])]
    Reasoning: Alter only the dolphin's color from blue to pink, without modifying other elements.
    Updated Objects: [('steam boat #1', [0.302, 0.293, 0.335, 0.194]), ('pink dolphin #1', [0.027, 0.324, 0.246, 0.160])]

- Example 4
    User prompt: Remove the leftmost bowl in this photo with two bowls and a dog.
    Current Objects: [('dog #1', [0.186, 0.592, 0.449, 0.408]), ('bowl #1', [0.376, 0.194, 0.324, 0.324]), ('bowl #2', [0.676, 0.494, 0.324, 0.324])]
    Reasoning: There are two bowls in the image and bowl #1 is identified as the leftmost one because its x coordinates (0.376) is smaller than that of bowl #2 (0.676).Thus, eliminate bowl #1 without modifying any remaining instances.
    Updated Objects: [('dog #1', [0.186, 0.592, 0.449, 0.408]), ('bowl #2', [0.676, 0.494, 0.324, 0.324])]

- Example 5
    User prompt: Add a pink bowl between two existing bowls in this photo.
    Current Objects: [('bowl #1', [0.076, 0.494, 0.324, 0.324]), ('bowl #2', [0.676, 0.494, 0.324, 0.324])]
    Reasoning: There are two bowls in the existing image. To add a pink bowl between the two, the x coordinates should be placed between 0.076 and 0.676 and the y coordinates should be between 0.494 and 0.494. When adding the object, be sure to prevent overlapping between existing objects and make sure the [top-left x-coordinate, top-left y-coordinate, top-left x-coordinate+box width, top-left y-coordinate+box height] lie between 0 and 1.
    Updated Objects: [('bowl #1', [0.076, 0.494, 0.324, 0.324]), ('bowl #2', [0.676, 0.494, 0.324, 0.324]), ('bowl #3', [0.376, 0.494, 0.324, 0.324])]

\end{lstlisting} \\\bottomrule
\end{tabular*}
\caption{Image editing in-context examples for the LLM controller.}
\label{tab:edit_example}
\end{table*}
}

\end{document}